\documentclass[runningheads]{llncs}
\usepackage{graphicx}
\usepackage{comment}
\usepackage{amsmath,amssymb}
\usepackage{color}

\usepackage[pagebackref=true,breaklinks=true,letterpaper=true,colorlinks,bookmarks=false]{hyperref}
\usepackage{enumitem, graphicx, subfigure, multirow, amsfonts, amsmath, xcolor, verbatim, capt-of}
\usepackage{algorithm, algorithmicx, algpseudocode}
\usepackage{booktabs, arydshln}

\newcommand{\eg}{\textit{e}.\textit{g}.}

\begin{document}

\pagestyle{headings}
\mainmatter
\def\ECCVSubNumber{4727} 

\title{Counterfactual Vision-and-Language Navigation via Adversarial Path Sampler}

\titlerunning{Fu et al.: Counterfactual Vision-and-Language Navigation} 
\authorrunning{Fu, Wang, Peterson, Grafton, Eckstein, and Wang} 
\author{Tsu-Jui Fu$^\dagger$, Xin Eric Wang$^\ddagger$, Matthew F. Peterson$^\dagger$,\\Scott T. Grafton$^\dagger$, Miguel P. Eckstein$^\dagger$, William Yang Wang$^\dagger$}
\institute{$^\dagger$UC Santa Barbara
\\
{\tt tsu-juifu@ucsb.edu, \{peterson, scott.grafton, miguel.eckstein\}@psych.ucsb.edu, william@cs.ucsb.edu}
\\
$^\ddagger$UC Santa Cruz
\\
{\tt xwang366@ucsc.edu}}

\maketitle

\begin{abstract}
Vision-and-Language Navigation (VLN) is a task where agents must decide how to move through a 3D environment to reach a goal by grounding natural language instructions to the visual surroundings.
One of the problems of the VLN task is data scarcity since it is difficult to collect enough navigation paths with human-annotated instructions for interactive environments.
In this paper, we explore the use of counterfactual thinking as a human-inspired data augmentation method that results in robust models. Counterfactual thinking is a concept that describes the human propensity to create possible alternatives to life events that have already occurred.
We propose an adversarial-driven counterfactual reasoning model that can consider effective conditions instead of low-quality augmented data. In particular, we present a model-agnostic adversarial path sampler (APS) that learns to sample challenging paths that force the navigator to improve based on the navigation performance. APS also serves to do pre-exploration of unseen environments to strengthen the model's ability to generalize. We evaluate the influence of APS on the performance of different VLN baseline models using the room-to-room dataset (R2R). The results show that the adversarial training process with our proposed APS benefits VLN models under both seen and unseen environments. And the pre-exploration process can further gain additional improvements under unseen environments.
\end{abstract}

\section{Introduction}
Vision-and-language navigation (VLN) \cite{anderson2018r2r,chen2019touchdown} is a complex task that requires an agent to understand  natural language, encode visual information from the surrounding environment, and associate critical visual features of the scene and appropriate actions with the instructions to achieve a specified goal (usually to move through a 3D environment to a target destination).

\begin{figure}[t]
    \centering
    
    \begin{minipage}{.49\textwidth}
        \includegraphics[width=\linewidth]{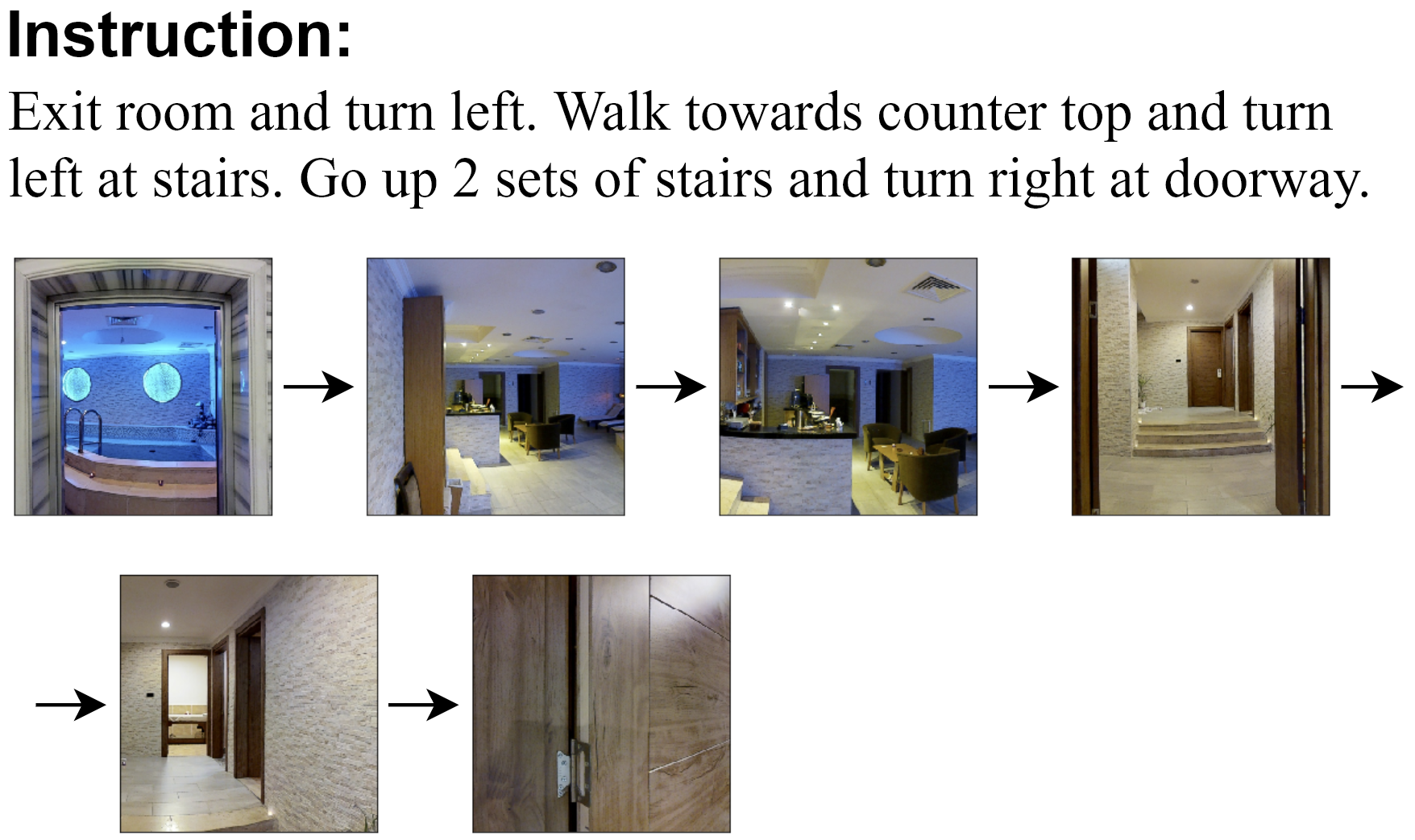}
    \end{minipage}~~
    \begin{minipage}{.49\textwidth}
        \includegraphics[width=\linewidth]{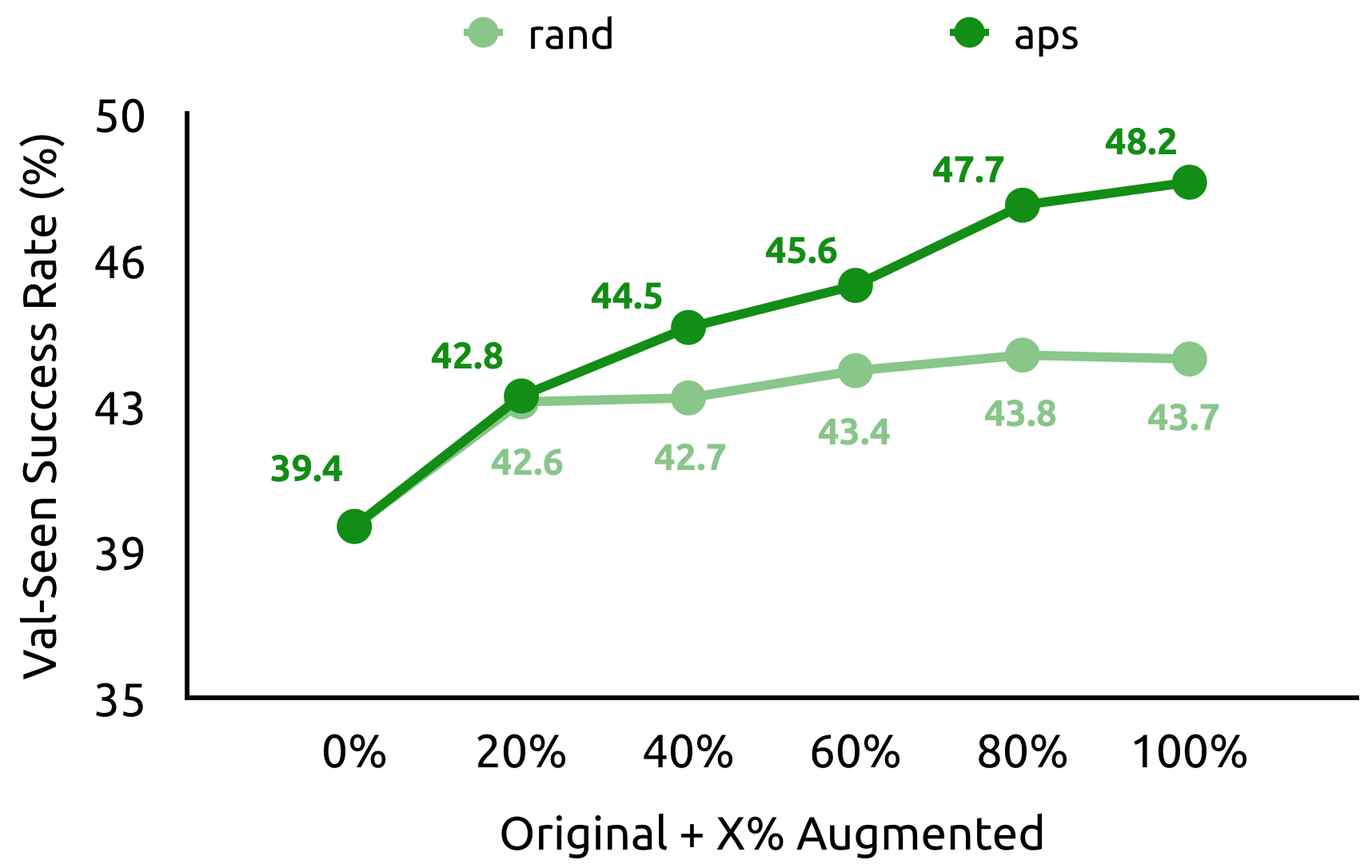} 
    \end{minipage}
    
    \vspace{-2ex}
    \caption{The comparison between randomly-sampled (rand) and APS-sampled (aps) under validation-seen set for Seq2Seq over different ratios of augmented path used.}
    \label{fig:pre-vs-aps}
    \vspace{-4ex}
\end{figure}

To accomplish the VLN task, the agent learns to align linguistic semantics and visual understanding and also make sense of dynamic changes in vision-and-language interactions. One of the primary challenges of the VLN task for artificial agents is data scarcity; for instance, while there are more than 200K possible paths in the Room-to-Room (R2R) dataset \cite{anderson2018r2r}, the R2R training data comprises only 14K sampled paths. This scarcity of data makes learning the optimal match between vision and language within the interactive environments quite challenging.

Meanwhile, humans often lack extensive experience with joint access to visual experience and accompanying language instructions for navigating novel or unfamiliar environments, yet the human mind can navigate environments despite this data scarcity by incorporating mechanisms such as counterfactual reasoning \cite{roese1997counterfactual} and self-recovered missing information. For example, if a human follows an instruction to ``turn right" and they see a door in front of them, they can also consider what they may have encountered had they turned left instead.  Or, if we stop in front of the dining table instead of walking away from it, what should the instruction be? The premise, then, is that counterfactual reasoning can improve performance in a VLN task through exploration and consideration of alternative actions that the agent did not actually make. This may allow the agent to operate in data-scarce scenarios by bootstrapping familiarity of environments and the links between instructions and multiple action policy options.

Counterfactual thinking has been used to increase the robustness of models for various tasks \cite{kusner2017cf-fair,garg2019cf-text}. However, no explicitly counterfactual models have been applied to the VLN task specifically. Speaker-Follower \cite{fried2018sf}, which applies a back-translated speaker model to reconstruct the instructions for randomly-sampled paths as augmented training examples, is probably the VLN model that comes closest to instantiating counterfactual thinking.

While the use of augmented training examples by the Speaker-Follower agent resembles a counterfactual process, the random sampling method is too arbitrary. Fig.~\ref{fig:pre-vs-aps} reports the performance of the model trained with randomly-sampled augmented data (the line in a light color) over different ratios of the augmented path used. It shows that the success rate stops increasing once augmented paths account for 60\% or more of the training data \cite{huang2019multi-dis-model}. Since those paths are all randomly sampled, it can limit the benefit of counterfactual thinking to data augmentation.

In this paper, we propose the use of adversarial-driven counterfactual thinking where the model learns to consider effective counterfactual conditions instead of sampling ample but uninformative data. We introduce a model-agnostic adversarial path sampler (APS) that learns how to generate augmented paths for training examples that are challenging, and thus effective, for the target navigation model. 
During the adversarial training process, the navigator is trying to accomplish augmented paths from APS and thus optimized for a better navigation policy, while the APS aims at producing increasingly challenging paths, which are therefore more effective than randomly-sampled paths.

Moreover, empowered by APS, the model can adapt to unseen environments in a practical setting---environment-based pre-exploration, where when deployed to a new environment, the robot can first pre-explore and get familiar with it, and then perform natural language guided tasks within this environment.

Experimental results on the R2R dataset show that the proposed APS can be integrated into a diverse collection of VLN models, improving their performance under both seen and unseen environments. In summary, our contributions are four-fold:
\begin{itemize}[noitemsep, topsep=1pt]
    \item We integrate counterfactual thinking into the vision-and-language navigation task, and propose the adversarial path sampler (APS) to progressively sample challenging and effective paths to improve the navigation policy.
    \item The proposed APS method is model-agnostic and can be easily integrated into various navigation models.
    \item Extensive experiments on the R2R dataset validate that the augmented paths generated by APS are not only useful in seen environments but also capable of generalizing the navigation policy better in unseen environments.  
    \item We demonstrate that APS can also be used to adapt the navigation policy to unseen environments under environment-based pre-exploration.
\end{itemize}

\section{Related Work}
\noindent\textbf{Vision-and-Language Navigation~}
Navigation in 3D environments based on natural language instruction has recently been investigated by many studies \cite{anderson2018r2r,chen2019touchdown,jain2019stay-path,ke2019tactical,ma2019regretful,huang2019vln-trans,wang2018rpa,ma2019self-m,fried2018sf,wang2019rcm,tan2019envdrop,hemachandra2015unseen,qi2020reverie,wang2020env-agn}. For vision-and-language navigation (VLN), fine-grained human-written instructions are provided as guidance to navigate a robot in indoor environments. But data scarcity is a critical issue in VLN due to the high cost of data collection.

In order to augment more data for training, the Speaker-Follower model \cite{fried2018sf} applies a back-translated speaker model to generate instructions for randomly-sampled paths. In spite of obtaining some improvements from those extra paths, a recent study \cite{huang2019multi-dis-model} shows that only a limited number of those augmented paths are useful and after using 60\% of the augmented data, the improvement diminishes with additional augmented data.  
In this paper, we present a model-agnostic adversarial path sampler that progressively produces more challenging paths via an adversarial learning process with the navigator, therefore forcing the navigation policy to be improved as the augmented data grows. \\

\noindent\textbf{Counterfactual Thinking~}
Counterfactual thinking is a concept that describes the human propensity to create possible alternatives to life events that have already occurred.
Humans routinely ask questions such as: ``What if ...?" or ``If there is only ..." to consider the outcomes of different scenarios and apply inferential reasoning to the process. In the field of data science, counterfactual thinking has been used to make trained models explainable and more robust \cite{kusner2017cf-fair,garg2019cf-text,goyal2019cf-vis}.
Furthermore, counterfactual thinking is also applied to augment training targets \cite{zmigrod2019cfd-text,chen2019cfc-sg,ashual2019cf-isg}. Although previous studies have shown some improvements over different tasks, they all implement counterfactual thinking arbitrarily without a selection process to sample counterfactual data that might optimize learning. This can limit the effectiveness of counterfactual thinking. In this paper, we combine the adversarial training with counterfactual conditions to guide models that might lead to robust learning. In this way, we can maximize the benefit of counterfactual thinking. \\

\noindent\textbf{Adversarial Training~}
Adversarial training refers to the process by which two models try to detrimentally influence each other's performance and as a result, both models improve by competing against each other.  Adversarial training has been successfully used to guide the target during model training \cite{goodfellow2014gan,wu2017adv-re,chou2018adv-pe,miyato2017adv-text,hong2019self-adv,agmon2017adv-robot}.
Apart from leading the training target, adversarial training is also applied to data augmentation \cite{antoniou2017gan-da,zhang2018meta-gan}. While previous studies just generate large amounts of augmented examples using a fixed pre-trained generator. In this paper, the generator is updated along with the target model and serves as a path sampler which samples challenging paths for effective data augmentation. \\

\noindent\textbf{Pre-Exploration under Unseen Environments~}
Pre-exploration under unseen environments is a popular method to bridge the gap between seen and unseen environments. Speaker-Follower \cite{fried2018sf} adopts a state-factored beam search for several candidate paths and then selects the best one. RCM \cite{wang2019rcm}  introduces self-imitation learning (SIL) that actively optimized the navigation model to maximize the cross-matching score between the generated path and the original instruction. Nevertheless, beam search requires multiple runs for each inference, and SIL utilizes the original instructions in the unseen environments for optimization. 
EnvDrop \cite{tan2019envdrop} conducts pre-exploration by sampling shortest paths from unseen environments and augments them with back-translation, which however utilizes the meta-information of unseen environments (\eg, the shortest path planner that the robot is not supposed to use). \\

\section{Methodology}
\subsection{Background}
\noindent\textbf{Visual-and-Language Navigation (VLN)~}
At each time step $t$, the environment presents the image scene $s_t$. After stepping an action $a_t$, the environment will transfer to next image scene $s_{t+1}$:
\begin{equation}
    s_{t+1} = \text{Environment}(s_t, a_t).
\end{equation}
To carry out a VLN task, the navigation model steps a serious of actions $\{a_t\}_{t=1}^{T}$ to achieve the final goal described in the instruction. Though previous studies propose different architectures of navigation model (NAV), in general, NAV is a recurrent action selector based on the visual feature of the image scene, navigation instruction, and previous history:
\begin{equation}
\begin{split}
    f_t &= \text{VisualFeature}(s_t), \\
    a_t &= \text{softmax}(\text{NAV}(f_t, I, h_t)),
\end{split}
\end{equation}
where $f_t$ is the visual feature of the image scene $s_t$  at time step $t$, $I$ is the navigation instruction, $h_t$ represents the previous history of image scenes, and $a_t$ is the probability of each action to step at time step $t$. 
With $a_t$, we can decide which action to step based on greedy decoding (step the action with the highest probability).

In this work, we experiment under navigator with 3 different architectures, Seq2Seq \cite{anderson2018r2r}, Speaker-Follower \cite{fried2018sf}, and RCM \cite{wang2019rcm}. \\

\noindent\textbf{Back-Translated Speaker Model~}
Introduced in Speaker-Follower \cite{fried2018sf}, the back-translated speaker model (Speaker) generates the instruction of a navigation path:
\begin{equation}
    I = \text{Speaker}(\{(f_1, a_1), (f_2, a_2), ..., (f_L, a_L)\}),
\end{equation}
where $f_t$ is the visual feature of the image scene, $a_t$ is the action taken at time step $t$, $L$ represents the length of the navigation path, and $I$ is the generated instruction. Speaker is trained with pairs of navigation paths and human-annotated instructions in the training data. With Speaker, we can sample various paths in the environments and augment their instructions.

\begin{figure}[t]
    \centering
    
    \begin{minipage}{.49\textwidth}
        \includegraphics[width=\linewidth]{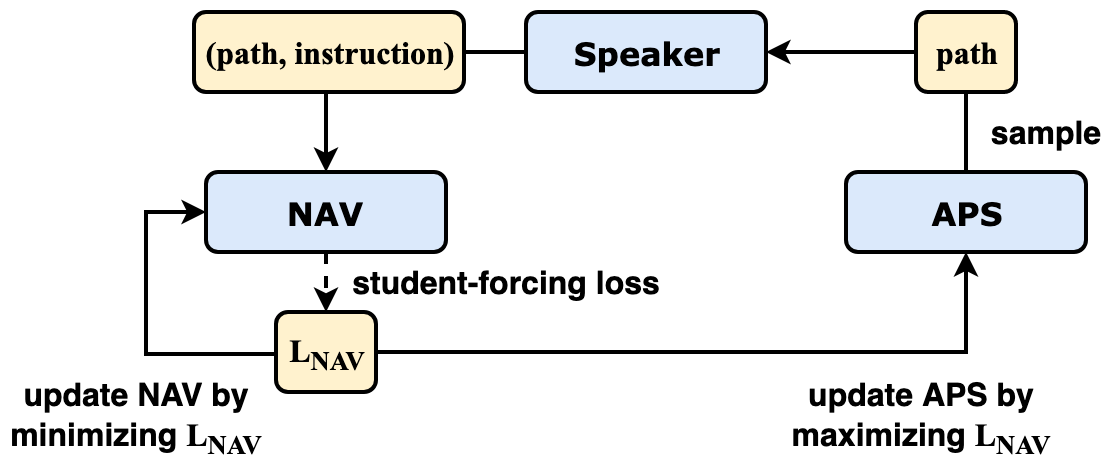}
        \vspace{-4ex}
        \caption{The learning framework of our adversarial path sampler (APS), where Speaker denotes the back-translated speaker model and NAV denotes the navigation model.}
        \label{fig:overview}
    \end{minipage}~~
    \begin{minipage}{.49\textwidth}
        \includegraphics[width=\linewidth]{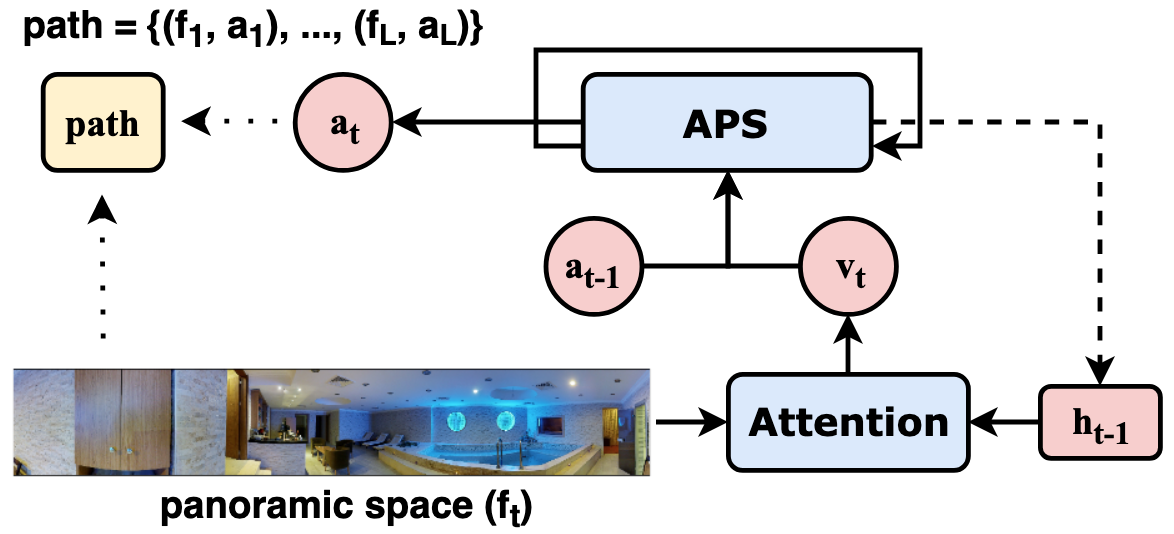} 
        \vspace{-4ex}
        \caption{The architecture of the adversarial path sampler (APS).}
        \label{fig:aps}
    \end{minipage}
    
    \vspace{-4ex}
\end{figure}

\subsection{Overview}
The overall learning framework of our model-agnostic adversarial path sampler (APS) is illustrated in Fig.~\ref{fig:overview}. At first, APS samples batch of paths $P$ and we adopt the Speaker \cite{fried2018sf}
to obtain the reconstructed instructions $I$. With the pairs of ($P$, $I$), we obtain the navigation loss $\mathcal{L}_{\text{NAV}}$. NAV minimizes $\mathcal{L}_{\text{NAV}}$ to improve navigation performance. While APS learns to sample paths that NAV can not perform so well by maximizing $\mathcal{L}_{\text{NAV}}$. Hence, there is an adversarial situation for $\mathcal{L}_{\text{NAV}}$ between NAV and APS, where APS aims at sampling challenging paths and NAV tries to solve the navigation tasks from APS.

By the above adversarial training process, we collect all of the ($P$, $I$) sampled from APS to compose the adversarial augmented data which can be more helpful to NAV than randomly-sampled one. Both Speaker and NAV are pre-trained using the original training set and Speaker keeps fixed during adversarial training. We collect all ($P$, $I$) sampled from APS as APS-sampled augmented path and further train NAV. More detail can be seen in Sec.~\ref{sec:train-aps}.

\subsection{Architecture of APS}
As shown in Fig.~\ref{fig:aps}, the proposed APS is a recurrent action sampler $\pi_\text{APS}$ which samples series of actions $\{a_t\}_{t=1}^{T}$ (with the scene images $\{f_t\}_{t=1}^{T}$ presented from the environment) and combines as the path output, where $f_t$ means the visual feature (e.g., extracted from the convolutional neural networks).
For the panoramic image scene, $f_{t, j}$ represents the visual feature of the image patch at viewpoint $j$ at time step $t$.

At each time step $t$, the history of previous visual feature and $a_{t-1}$ is encoded as $h_{t}$ by a long short-term memory (LSTM) \cite{hochreiter1997lstm} encoder:
\begin{equation}
    h_t = \text{LSTM}([v_t, a_{t-1}], h_{t-1}),
\end{equation}
where $a_{t-1}$ is the action taken at previous step and $v_t$ is the weighted sum of visual feature of each image path for the panoramic image scene. $v_t$ is calculated using the attention \cite{bahdanau2015att} between the history $h_{t-1}$ and the image patches $\{f_{t,j}\}_{j=1}^m$:
\begin{equation}
\begin{split}
    v_t &= \text{Attention}(h_{t-1}, \{f_{t, j}\}_{j=1}^{m}) \\
    &=\sum_{j} \text{softmax}(h_{t-1}W_{h}(f_{t, j}W_{f})^{T})f_{t, j},
\end{split}
\end{equation}
where $W_h$ and $W_f$ are learnable projection matrics. The above equation of $v_t$ is for panoramic scene with $m$ viewpoints. APS also supports the navigator which uses visuomotor view as input (e.g., Seq2Seq \cite{anderson2018r2r}) and the single visual feature $f_t$ is seen as $v_t$ directly.

Finally, APS decides which action to step based on the history $h_t$ and action embedding $u$:
\begin{equation}
    a_t = \text{softmax}(h_{t}W_{c}(u_{k}W_{u}^{T})),
\end{equation}
where $u_{k}$ is the action embedding of the $k$-th navigable direction. $W_c$ and $W_u$ are learnable projection matrics.

\begin{algorithm}[t]
    \begin{algorithmic}[1]
        \small
    
        \State NAV: the target navigation model
        \State Speaker: the back-translated instruction model
        \State APS: the adversarial path sampler
        \State $\text{aug}_{\text{aps}}$: collected APS-sampled augmented data
        \\
        \State Pre-train NAV with original training set
        \State Pre-train Speaker with original navigation path
        \State Initialize APS
        \State $\text{aug}_{\text{aps}}$ $\leftarrow$ $\varnothing$
        \\
        \While{DO\_APS}
            \State $P = \{(f_1, a_1), (f_2, a_2), ..., (f_L, a_L)\}$ $\leftarrow$ APS samples
            \State $I$ $\leftarrow$ back-translated by Speaker with $P$
            \State $\mathcal{L}_\text{NAV}$ $\leftarrow$ student-forcing loss of NAV using ($P$, $I$)
            \\
            \State Update NAV by minimizing $\mathcal{L}_\text{NAV}$
            \State Update APS by maximizing $\mathcal{L}_\text{NAV}$ using Policy Gradient
            \State $\text{aug}_{\text{aps}}$ $\leftarrow$ $\text{aug}_{\text{aps}}$ $\cup$ ($P$, $I$)
        \EndWhile
        \\
        \State Train NAV with $\text{aug}_\text{aps}$
        \State Fine-tune NAV with original training set 
    \end{algorithmic}

    \caption{Training Process of Adversarial Path Sampler}
    \label{algo:aps}
\end{algorithm}

\subsection{Adversarial Training of APS} \label{sec:train-aps}
After each unrolling of APS, we comprise the navigation history $\{a_t\}_{t=1}^{T}$ and $\{f_{t, j}\}_{j=1}^{m}$ to obtain the path $P$. 
To be consistent with the original training data whose navigation paths are all shortest paths \cite{anderson2018r2r}, we transform the sampled paths by APS into shortest paths\footnote{Note that transforming the sampled paths into shortest paths can only be done under seen environments. For pre-exploration under unseen environments, we directly use the sampled paths because the shortest path planner should not be exploited in unseen environments.} (same start and end nodes as in the sampled paths). 
Then we employ the Speaker model~\cite{fried2018sf} to produce one instruction $I$ for each sampled path $P$, and eventually obtain a set of new augmented pairs ($P$, $I$). 
We train the navigation model (NAV) with ($P$, $I$) using student-forcing \cite{anderson2018r2r}. The training loss ($\mathcal{L}_{\text{NAV}}$)
can be seen as an indicator of NAV's performance under ($P$, $I$): the higher $\mathcal{L}_{\text{NAV}}$ is, the worse NAV performs. Hence, in order to create increasingly challenging paths to improve the navigation policy, we define the loss function $\mathcal{L}_\text{APS}$ of APS as:
\begin{equation}
    \mathcal{L}_\text{APS} = - \mathbb{E}_{p(\text{P};\pi_\text{APS})}\mathcal{L}_{\text{NAV}}.
\end{equation}

Since the path sampling process is not differentiable, we adopt policy gradient \cite{sutton2000pg} and view $\mathcal{L}_{\text{NAV}}$ as the reward $R$ to optimize the APS objective. According to the REINFORCE algorithm \cite{williams1992reinforce}, the gradient is computed as following:
\begin{equation}
\begin{aligned}
    \nabla_{\pi_\text{APS}} \mathcal{L}_\text{APS} &\approx -\sum_{t=1}^{T} [\nabla_{\pi_\text{APS}} \log p(a_t | a_{1:t-1} ; \pi_\text{APS}) R] \\
    &\approx -\sum_{t=1}^{T} [\nabla_{\pi_\text{APS}} \log p(a_t | a_{1:t-1} ; \pi_\text{APS}) (R-b)],
\end{aligned}
\end{equation}
where $b$ is the baseline estimation to reduce the variance and we treat $b$ as the mean of all previous losses. Note that APS is model-agnostic and can be easily integrated into different navigation models, since it only considers the training loss from a navigation model regardless of its model architecture.

Algorithm~\ref{algo:aps} illustrates the training process of APS. APS aims at maximizing the navigation loss $\mathcal{L}_{\text{NAV}}$ of NAV to create more challenging paths, while NAV tries to minimize $\mathcal{L}_{\text{NAV}}$ to do better navigation:  
\begin{equation}
\begin{split}
    \min_{\text{NAV}} & \max_{\text{APS}} \mathcal{L}_{\text{NAV}}.
\end{split}
\end{equation}
After collecting the challenging paths augmented by APS, we train NAV on them and finally fine-tune NAV with the original training set. The detailed analysis of APS-sampled augmented data is shown in Sec.~\ref{sec:ayz-aps}.

\begin{figure}[t]
    \centering
    
    \includegraphics[width=.5\linewidth]{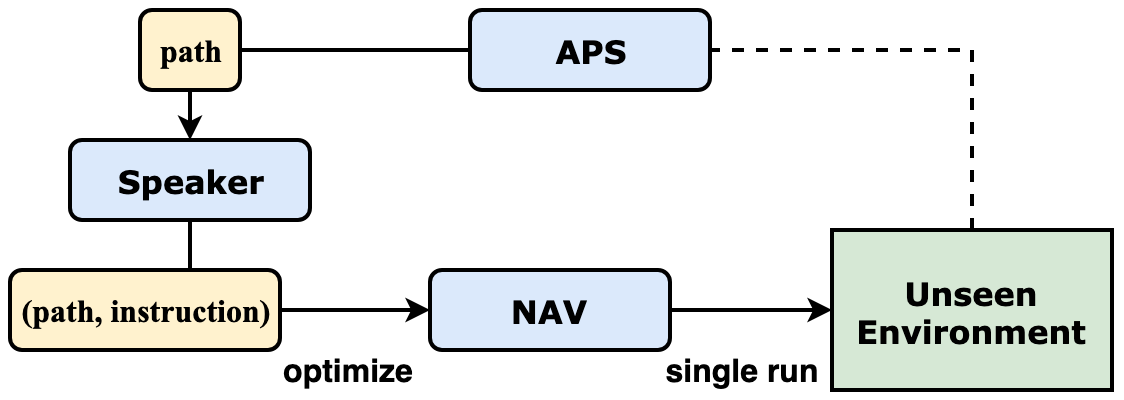}
    \vspace{-2ex}
    \caption{The optimization flow of environment-based pre-exploration under unseen environments. APS samples paths from the unseen environment to optimize NAV and make it more adaptive. Then, NAV runs each instruction in a single turn.}
    \label{fig:pre-exp}
    
    \vspace{-4ex}
\end{figure}

\subsection{Environment-based Pre-Exploration}
Pre-exploration is a technique that adapts the navigation model to unseen environments. The navigator can explore the unfamiliar environment first and increase the chance to carry out the navigation instructions under unseen environments.
For previous pre-exploration methods like beam search \cite{fried2018sf} or self-imitation learning (SIL) \cite{wang2019rcm}, they are instruction-based which optimizes for each instruction. This will make the navigation path excessive long since it first runs many different paths and then selects the possible best one. 

In the real world, when we deploy a robot into a new environment, it might pre-explore and get familiar with the environment, and then efficiently execute the tasks following natural language instructions within this environment. So unlike previous approaches~\cite{wang2019rcm,tan2019envdrop} that either optimize the given instructions or assume access to all the unseen environments at once, we propose to use our APS method to do the environment-based pre-exploration where the agent pre-explore an environment only for the tasks within the same environment with no prior knowledge of it. Under an unseen environment, we adopt APS to sample multiple paths ($P'$) and generate the instructions ($I'$) of the sampled paths\footnote{Note that the shortest-path information is not used during pre-exploration.} with the Speaker model \cite{fried2018sf}. We then use ($P'$, $I'$) to optimize NAV to adapt to the unseen environment as illustrated in Fig.~\ref{fig:pre-exp}. 
Note that during pre-exploration, we only optimize NAV and let APS fixed\footnote{We have tried to update APS simultaneously with NAV during pre-exploration, but it turns out that under a previous unseen environment without any regularization of human-annotated paths, APS tends to sample too difficult paths to accomplish, \eg, back and forth or cycles. However, those paths will not improve NAV and may even hurt the performance. To avoid this kind of dilemma, we keep APS fixed under the pre-exploration.}. We also present a detailed analysis of our proposed environment-based pre-exploration method in Sec.~\ref{sec:ayz-pre-exp}.

\begin{table}[t]
    \centering
    
    \scriptsize
    \setlength{\tabcolsep}{5pt}
    \begin{tabular}[t]{lcccc}
        \toprule
        ~ & \multicolumn{4}{c}{\textbf{Test} (VLN Challenge Leaderboard)} \\
        \cmidrule{2-5}
        Model & NE $\downarrow$ & OSR $\uparrow$ & SR $\uparrow$ & SPL $\uparrow$ \\
        \midrule
        Seq2Seq \cite{anderson2018r2r} & 7.9 & 26.6 & 20.4 & 18.0 \\
        ~+ $\text{aug}_\text{rand}$ & 7.8 & 26.2 & 21.0 & 18.8 \\
        ~+ $\text{aug}_\text{aps}$ & \textbf{7.5} & \textbf{30.1} & \textbf{22.5} & \textbf{19.3} \\
        \hdashline
        ~+ $\text{aug}_\text{aps}$+pre-exploration & \textbf{6.7} & 29.4 & \textbf{23.2} & \textbf{20.8} \\
        \midrule
        Speaker-Follower \cite{fried2018sf} & 7.0 & 41.2 & 30.9 & 24.0 \\
        ~+ $\text{aug}_\text{rand}$ & 6.6 & 43.4 & 34.8 & 29.2 \\
        ~+ $\text{aug}_\text{aps}$ & \textbf{6.5} & \textbf{44.2} & \textbf{36.1} & 28.8 \\
        \hdashline
        ~+ $\text{aug}_\text{aps}$+pre-exploration & \textbf{5.9} & \textbf{46.4} & \textbf{37.6} & \textbf{32.4} \\
        \midrule
        RCM \cite{wang2019rcm} & 6.7 & 43.5 & 35.9 & 33.1 \\
        ~+ $\text{aug}_\text{rand}$ & 5.9 & 52.4 & 44.5 & 40.8 \\
        ~+ $\text{aug}_\text{aps}$ & \textbf{5.8} & \textbf{53.9} & \textbf{45.1} & \textbf{40.9} \\
        \hdashline
        ~+ $\text{aug}_\text{aps}$+pre-exploration & \textbf{5.5} & \textbf{55.6} & \textbf{45.9} & \textbf{40.9} \\
        \bottomrule
        \\
    \end{tabular}
    \caption{R2R results for Seq2Seq, Speaker-Follower, and RCM under testing set. Models are trained without augmented data, with randomly-sampled augmented path ($\text{aug}_{\text{rand}}$), with APS-sampled augmented path ($\text{aug}_{\text{aps}}$), and under pre-exploration in unseen environments. Note that those results are run in single turn and with greedy action decoding.}
    \label{table:result-test}
    
    \vspace{-4ex}
\end{table}

\begin{table}[t]
    \centering
    
    \scriptsize
    \setlength{\tabcolsep}{5pt}
    \begin{tabular}[t]{lccccccccc}
        \toprule
        ~ & \multicolumn{4}{c}{\textbf{Val-Seen}} & ~ & \multicolumn{4}{c}{\textbf{Val-Unseen}} \\
        \cmidrule{2-5} \cmidrule{7-10}
        Model & NE $\downarrow$ & OSR $\uparrow$ & SR $\uparrow$ & SPL $\uparrow$ & ~ & NE $\downarrow$ & OSR $\uparrow$ & SR $\uparrow$ & SPL $\uparrow$ \\
        \midrule
        Seq2Seq \cite{anderson2018r2r} & 6.0 & 51.7 & 39.4 & 33.8 & ~ & 7.8 & 27.7 & 22.1 & 19.1 \\
        ~+ $\text{aug}_\text{rand}$ & 5.3 & 58.1 & 43.7 & 37.2 & ~ & 7.7 & 28.9 & 22.6 & 19.9 \\
        ~+ $\text{aug}_\text{aps}$ & \textbf{5.0} & \textbf{60.8} & \textbf{48.2} & \textbf{40.1} & ~ & \textbf{7.1} & \textbf{32.7} & \textbf{24.2} & \textbf{20.4} \\
        \hdashline
        ~+ $\text{aug}_\text{aps}$+pre-exploration & \multicolumn{4}{c}{-} & ~ & \textbf{6.6} & \textbf{37.8} & \textbf{27.0} & \textbf{24.6} \\
        \midrule
        Speaker-Follower \cite{fried2018sf} & 5.0 & 61.6 & 51.7 & 44.4 & ~ & 6.9 & 40.7 & 29.9 & 21.0 \\
        ~+ $\text{aug}_\text{rand}$ & 3.7 & 74.2 & 66.4 & 59.8 & ~ & 6.6 & 46.6 & 36.1 & 28.8 \\
        ~+ $\text{aug}_\text{aps}$ & \textbf{3.3} & \textbf{74.9} & \textbf{68.2} & \textbf{62.5} & ~ & \textbf{6.1} & \textbf{46.7} & \textbf{38.8} & \textbf{32.1} \\
        \hdashline
        ~+ $\text{aug}_\text{aps}$+pre-exploration & \multicolumn{4}{c}{-} & ~ & \textbf{5.2} & \textbf{49.1} & \textbf{42.0} & \textbf{35.7} \\
        \midrule
        RCM \cite{wang2019rcm} & 5.7 & 53.8 & 47.0 & 44.3 & ~ & 6.8 & 43.0 & 35.0 & 31.4 \\
        ~+ $\text{aug}_\text{rand}$ & 4.1 & 66.9 & 61.9 & 58.6 & ~ & 5.7 & 52.4 & 45.6 & 41.8 \\
        ~+ $\text{aug}_\text{aps}$ & \textbf{3.9} & \textbf{69.3} & \textbf{63.2} & \textbf{59.5} & ~ & \textbf{5.4} & \textbf{56.6} & \textbf{47.7} & \textbf{42.8} \\
        \hdashline
        ~+ $\text{aug}_\text{aps}$+pre-exploration & \multicolumn{4}{c}{-} & ~ & \textbf{5.3} & 56.2 & \textbf{48.0} & \textbf{42.8} \\
        \bottomrule
        \\
    \end{tabular}
    \caption{R2R results for Seq2Seq, Speaker-Follower, and RCM under validation-seen and validation-unseen sets. Models are trained without augmented data, with randomly-sampled augmented path ($\text{aug}_{\text{rand}}$), with APS-sampled augmented path ($\text{aug}_{\text{aps}}$), and under pre-exploration in unseen environments. Note that those results are run in single turn and with greedy action decoding.}
    \label{table:result}
    
    \vspace{-4ex}
\end{table}

\section{Experiments}
\subsection{Experimental Setup}
\noindent\textbf{R2R Dataset~}
We evaluate the proposed method on the Room-to-Room (R2R) dataset \cite{anderson2018r2r} for vision-and-language navigation. R2R is built upon the Matterport3D \cite{angel2017mp3d}, which contains 90 different environments that are split into 61 for training and validation-seen, 11 for validation-unseen, and 18 for testing sets. There are 7,189 paths and each path has 3 human-written instructions. The validation-seen set shares the same environments with the training set. In contrast, both the validation-unseen and the testing sets contain distinct environments that do not appear during training. \\

\noindent\textbf{Evaluation Metrics~}
To compare with the existing methods, we report the same used evaluation metrics: Navigation Error (NE), Oracle Success Rate (OSR), Success Rate (SR), and Success Rate weighted by Path Length (SPL). NE is the distance between the agent's final position and goal location. OSR is the success rate at the closest point to the goal that the agent has visited. SR is calculated as the percentage of the final position within 3m from the goal location. SPL, defined in \cite{anderson2018spl}, is the success rate weighted by path length which considers both effectiveness and efficiency. \\

\noindent\textbf{Baselines~}
We experiment with the effectiveness of the model-agnostic APS on 3 kinds of baselines:
\begin{itemize}[noitemsep, topsep=2pt, leftmargin=0.5cm]
    \item \textbf{Seq2Seq} \cite{anderson2018r2r}, the attention-based seq2seq model that is trained with student forcing (or imitation learning) under the visuomotor view and action space; 
    \item \textbf{Speaker-Follower} \cite{fried2018sf}, the compositional model that is trained with student forcing under the panoramic view and action space;
    \item \textbf{RCM} \cite{wang2019rcm}, the model that integrates cross-modal matching loss, and is trained using reinforcement learning under the panoramic view and action space.
\end{itemize}
In the following sections, we use the notations as:
\begin{itemize}[noitemsep, topsep=2pt, leftmargin=0.5cm]
    \item $\text{aug}_\text{rand}$: the randomly-sampled augmented path;
    \item $\text{aug}_\text{aps}$: the APS-sampled augmented path;
    \item $\text{model}_\text{rand}$: the model trained with $\text{aug}_\text{rand}$;
    \item $\text{model}_\text{aps}$: the model trained with $\text{aug}_\text{aps}$.
\end{itemize}
For example, $\text{Speaker-Follower}_\text{aps}$ is the Speaker-Follower model trained with the APS-sampled augmented path.

For each baseline, we report the results of the model trained without any augmented data, trained with $\text{aug}_\text{rand}$, and trained with $\text{aug}_\text{aps}$. For the unseen environments, we also report the results under the pre-exploration. \\

\noindent\textbf{Implementation Details~}
To follow the previous studies \cite{anderson2018r2r,fried2018sf,wang2019rcm}, we adopt ResNet-152 \cite{he2016res-net} to extract visual features (2048d) for all scene images without fine-tuning; for the navigation instructions, the pre-trained GloVe embeddings \cite{pennington2014glove} are used for initialization and then fine-tuned with the model training. For baseline models, we apply the same batch size 100, LSTM with 512 hidden units, learning rate 1e-4, RL learning rate 1e-5, and dropout rate 0.5. For our proposed APS, the hidden unit of LSTM is also 512, the action embedding size is 128, and the learning rate is 3e-5. We adopt the learning rate 1e-5 under the pre-exploration for the unseen environments. All models are optimized via Adam optimizer \cite{kingma2015adam} with weight decay 5e-4.

For $\text{aug}_\text{rand}$, we use the same 17K paths as Speaker-Follower \cite{fried2018sf}. To compare fairly, APS also adversarially samples the same amounts of paths for data augmentation. The navigation models are first trained using augmented data for 50K iterations and then fine-tuned with original human-written instructions for 20K iterations.

\begin{figure}[t]
    \centering
    
    \includegraphics[width=\linewidth]{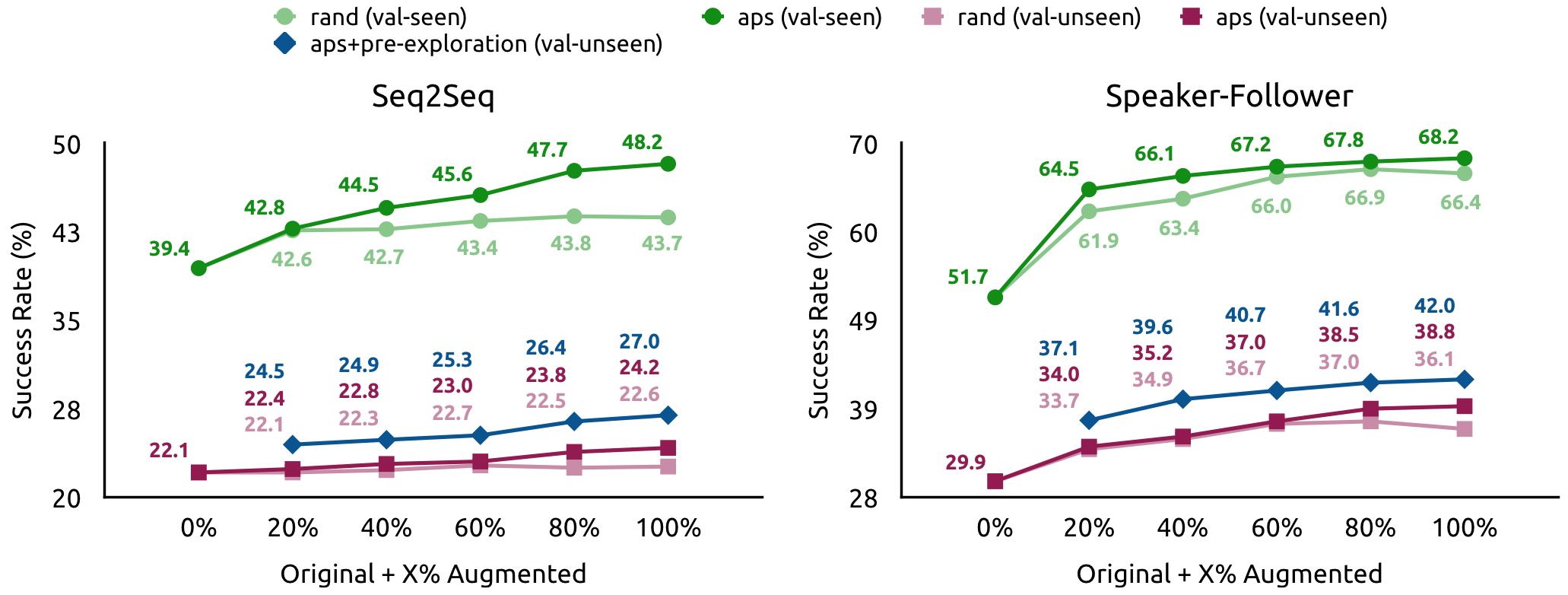}
    \vspace{-4ex}
    \caption{The comparison between randomly-sampled and APS-sampled under validation-seen and validation-unseen sets for Seq2Seq and Speaker-Follower over different ratios of augmented path used.}
    \label{fig:aps-s2s-sf}
    
    \vspace{-4ex}
\end{figure}

\subsection{Quantitative Results}
Table.~\ref{table:result-test} and \ref{table:result} present the R2R results for Seq2Seq~\cite{anderson2018r2r}, Speaker-Follower~\cite{fried2018sf}, and RCM~\cite{wang2019rcm} under validation-seen, validation-unseen, and testing sets. All models are trained without augmented data, with $\text{aug}_\text{rand}$, and with $\text{aug}_\text{aps}$. 
First, we can observe that under validation-seen set, $\text{Seq2Seq}_\text{aps}$ outperforms $\text{Seq2Seq}_\text{rand}$ on all evaluation metrics, \eg, 4.5\% absolute improvement on Sucess Rate and 2.9\% on SPL. Similar trends can be found for Speaker-Follower and RCM where models trained with APS-sampled paths comprehensively surpass models trained with randomly-sampled paths. Since APS can sample increasingly challenging and custom-made paths for the navigator, APS-sampled paths are more effective than randomly-sample paths and bring in larger improvements on all metrics for all navigation models.

For the unseen environments, all models trained with APS consistently outperform $\text{model}_\text{rand}$ with 1.6\%-2.7\% success rate under validation-unseen set and 0.6\%-1.5\% under testing set. The improvement shows that APS-sampled paths are not only helpful under the seen environments, but also strengthens the model's generalizability under the unseen environments. The results under validation-seen, validation-unseen, and testing sets demonstrate that our proposed APS can further improve the baseline models in all terms of visuomotor view, panoramic view, imitation learning, and reinforcement learning.

And under the pre-exploration, all models gain further improvement, especially on SPL for Seq2Seq and Speaker-Follower due to the prior exploration experience which can shorten the navigation path length. For RCM, they adopt reinforcement learning which may increase the path length but still obtain improvement on success rate.

\begin{figure}[t]
    \centering
    
    \begin{minipage}{.49\textwidth}
        \centering
        \small
        
        \begin{tabular}[t]{cccc}
            \toprule
            ~ & \multicolumn{3}{c}{As Testing Set} \\
            \cmidrule{2-4}
            Model & train & $\text{aug}_{\text{rand}}$ & $\text{aug}_{\text{aps}}$ \\
            \midrule
            Seq2Seq & 71.3 & 20.3 & 17.7 \\
            $\text{Seq2Seq}_\text{rand}$ & \textbf{81.4} & 26.4 & 23.8 \\
            $\text{Seq2Seq}_\text{aps}$ & 78.5 & \textbf{27.3} & \textbf{24.8} \\
            \bottomrule
        \end{tabular}
    \end{minipage}~~
    \begin{minipage}{.49\textwidth}
        \centering
        \small
        
        \begin{tabular}[t]{ccc}
            \toprule
            ~ & \multicolumn{2}{c}{As Testing Set} \\
            \cmidrule{2-3}
            Model & $\text{aug}_{\text{rand}}$ & $\text{aug}_{\text{aps}}$ \\
            \midrule
            $\text{RCM}_\text{rand}$ & 33.3 & 31.1 \\
            $\text{RCM}_\text{aps}$ & \textbf{38.9} & \textbf{37.9} \\
            \bottomrule
        \end{tabular}
    \end{minipage}
    
    \caption{The success rate under training, randomly-sampled augmented ($\text{aug}_{\text{rand}}$), and APS-sampled augmented ($\text{aug}_{\text{aps}}$) sets for Seq2Seq and RCM.}
    \label{table:diff-pre-aps}
    
    \vspace{-4ex}
\end{figure}

\subsection{Ablation Study}
\noindent\textbf{Random Path Sampling vs Adversarial Path Sampling~} \label{sec:ayz-aps}
To investigate the advantage of APS, we perform a detailed comparison between randomly-sampled and APS-sampled data. Fig.~\ref{fig:aps-s2s-sf} presents the R2R success rate over different ratios of augmented data used for Seq2Seq and Speaker-Follower. 
The trend line in light color shows that $\text{Seq2Seq}_\text{rand}$ cannot gain additional improvement when using more than 60\% augmented data. However, for our proposed APS, the sampled augmented path can keep benefiting the model when more data used and achieve 4.5\% and 1.6\% improvement under validation-seen and validation-unseen sets, respectively. 
Since $\text{aug}_\text{rand}$ is sampled in advance, the help to the model is limited. While on the other hand, our proposed APS adversarially learns to sample challenging paths that force the navigator to keep improving. A similar trend can be found for Speaker-Follower where the improvement of $\text{Speaker-Follower}_\text{rand}$ is also stuck but $\text{Speaker-Follower}_\text{aps}$ can lead to even better performance. \\

\noindent\textbf{Difficulty and Usefulness of the APS-sampled Paths~}
For a more intuitive view of the difficulty and usefulness of the APS-sampled paths, we conduct experiments shown in Table~\ref{table:diff-pre-aps} to quantitatively compare them with randomly-sample paths.
As you can see, the APS-sampled paths seem to be the most challenging as all models perform worst on them. These paths can in turn help train a more robust navigation model ($\text{Seq2Seq}_\text{aps}$) that outperforms the model trained with randomly sampled paths. Moreover, $\text{Seq2Seq}_{\text{aps}}$ even performs better on $\text{aug}_\text{rand}$ than $\text{Seq2Seq}_{\text{rand}}$ which shows that $\text{aug}_\text{aps}$ is not only challenging but also covers useful paths over $\text{aug}_\text{rand}$. \\

\begin{figure}[t]
    \centering
    
    \begin{minipage}{.49\textwidth}
        \includegraphics[width=\linewidth]{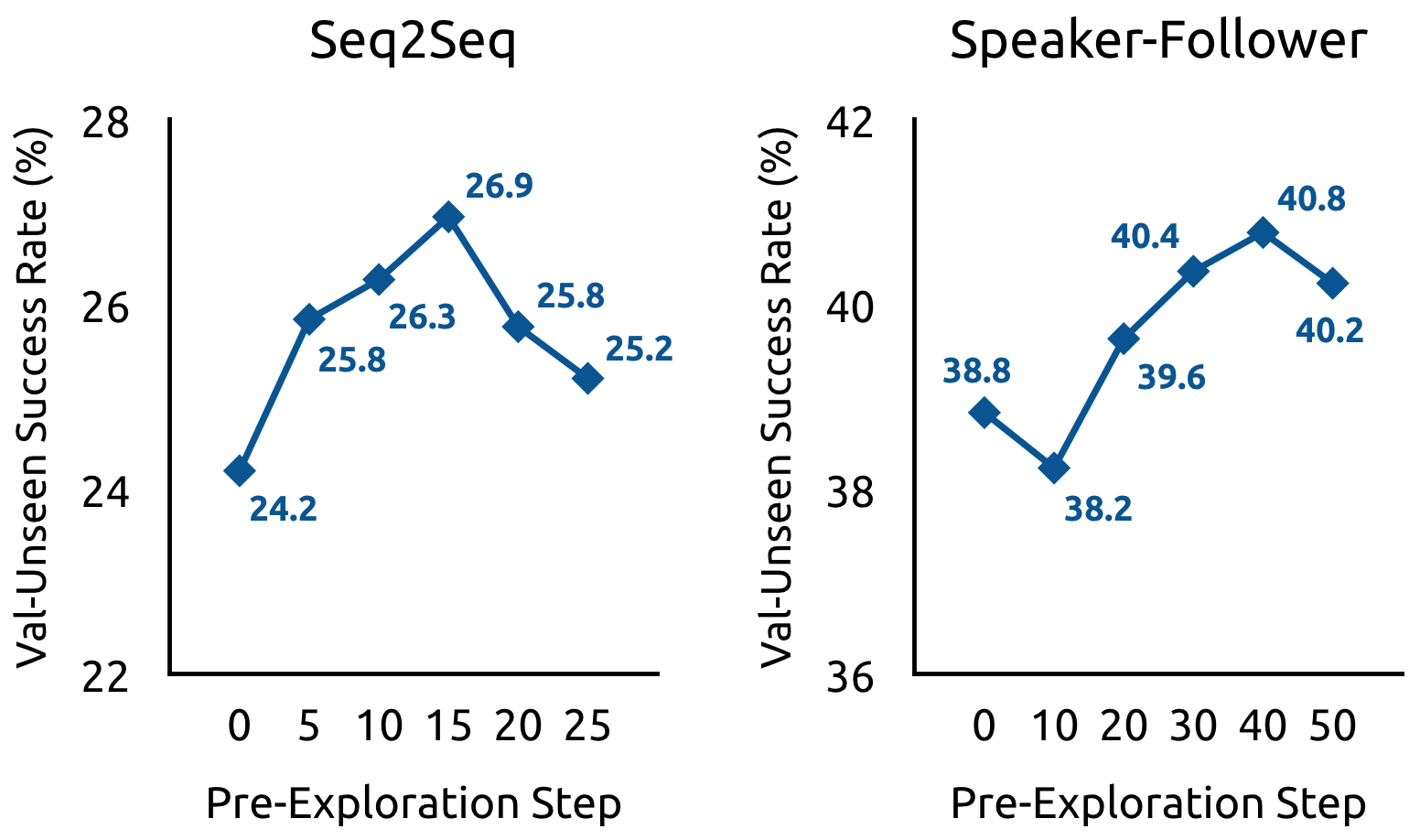}
        \vspace{-4ex}
        \caption{The success rate under validation-unseen set under different pre-exploration steps for Seq2Seq and Speaker-Follower.}
        \label{fig:pre-exp-s2s-sf}
    \end{minipage}~~
    \begin{minipage}{.49\textwidth}
        \includegraphics[width=\linewidth]{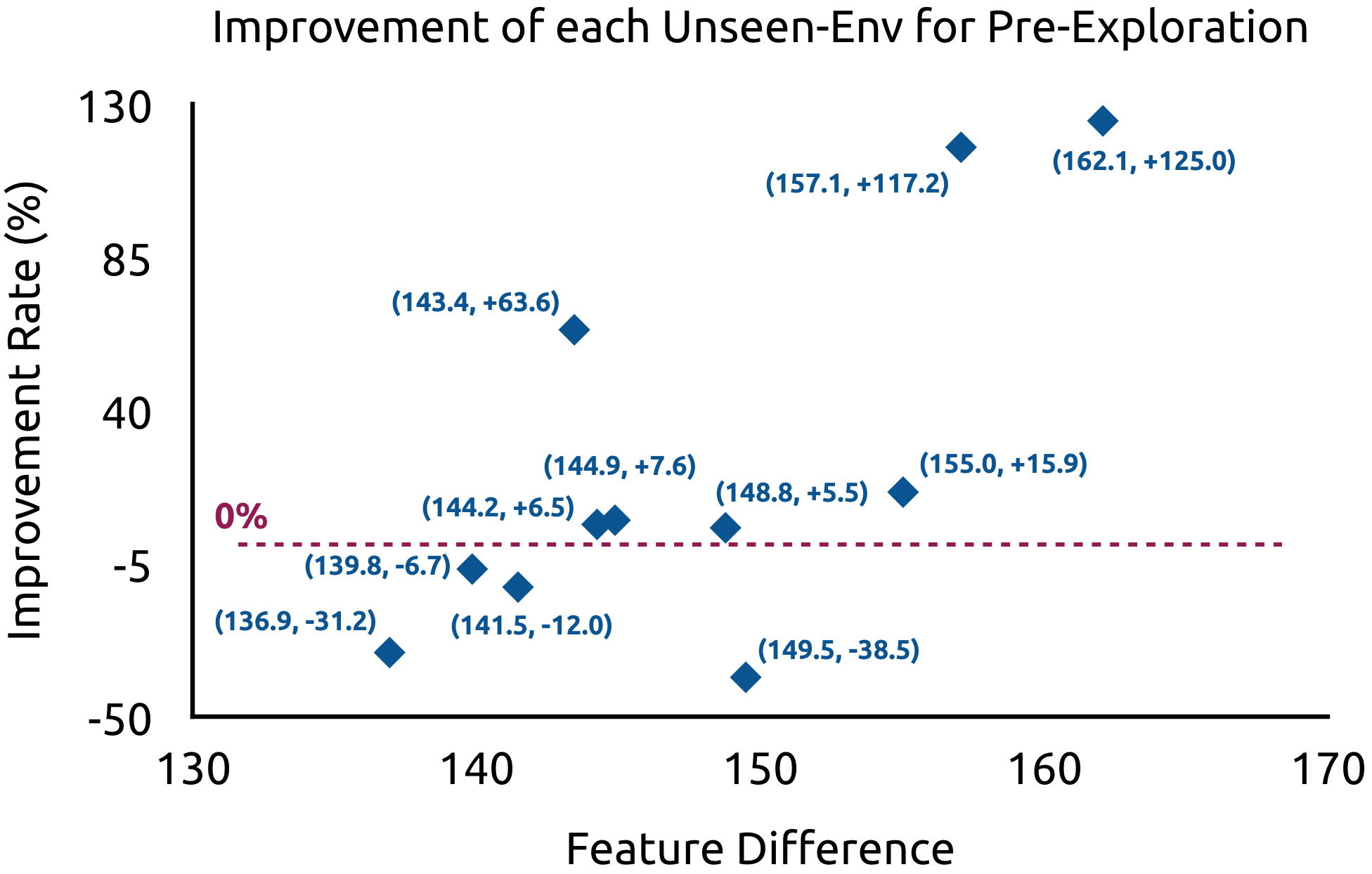}
        \vspace{-4ex}
        \caption{The improvement of success rate over the scene feature difference under each validation-unseen environment under the pre-exploration. Each point represents a distinct validation-unseen environment.}
        \label{fig:pre-exp-each}
    \end{minipage}
    
    \vspace{-4ex}
\end{figure}

\begin{figure}[t]
    \centering
    
    \includegraphics[width=\linewidth]{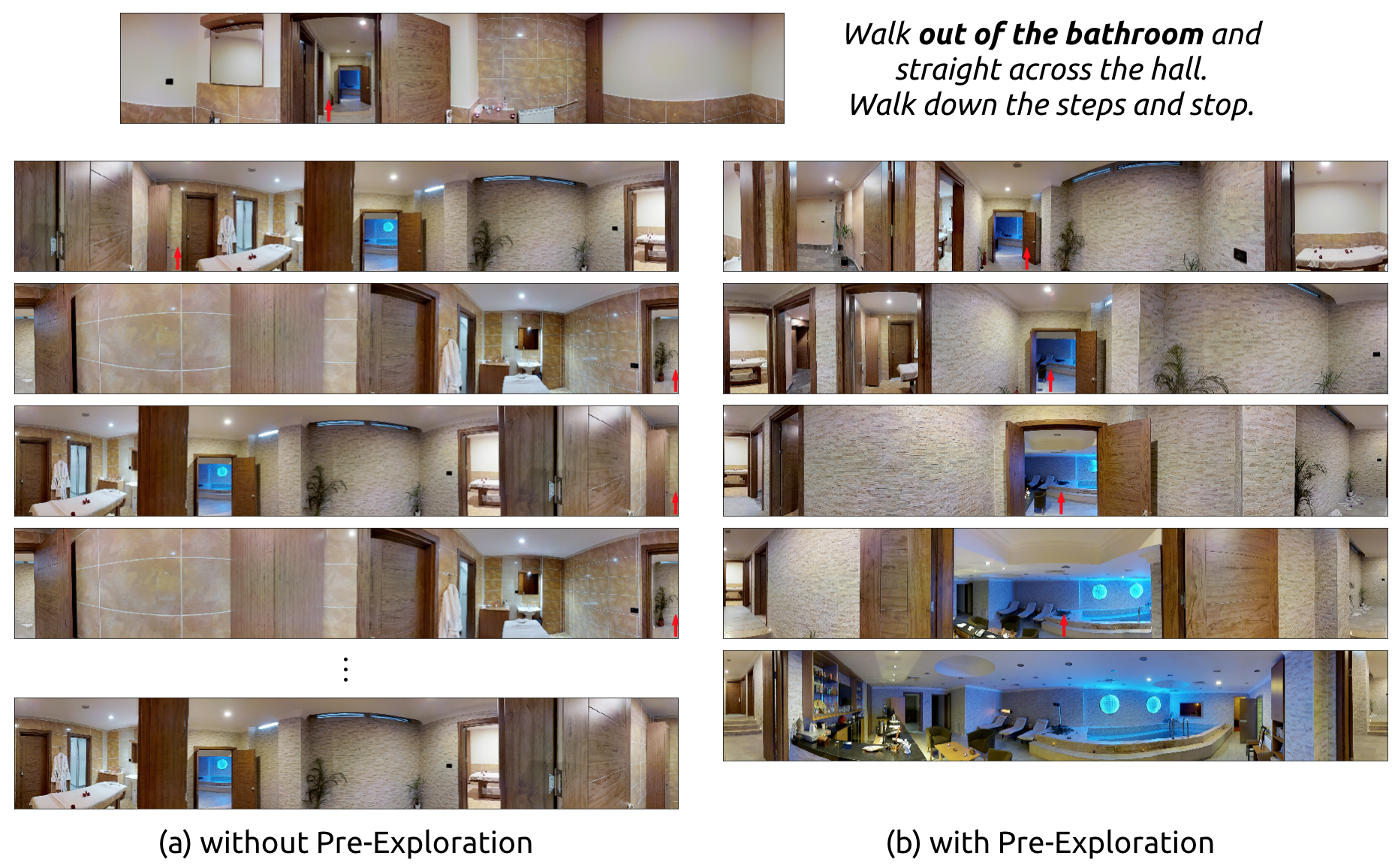}
    \vspace{-4ex}
    \caption{The visualization example of the comparison between without and with the pre-exploration under the validation-unseen environment.}
    \label{fig:vis}
    
    \vspace{-4ex}
\end{figure}

\noindent\textbf{Pre-Exploration~} \label{sec:ayz-pre-exp}
Table.~\ref{table:result} has shown the improvement brought from the pre-exploration. While, those paths in training, validation, and testing sets are all shortest path but the paths sampled from our APS under unseen environments are not promised to be the shortest. With more pre-exploration steps, the model has more opportunities to explore the unseen environment but at the same time, those too complicated paths sampled from APS may hurt the model. Fig.~\ref{fig:pre-exp-s2s-sf} presents the success rate under different pre-exploration steps. It shows a trade-off between the model performance and the iterations of the pre-exploration. For Seq2Seq, 15 steps of pre-exploration come out the best result and 40 steps are most suitable for Speaker-Follower.

We also analyze the performance under the pre-exploration under each unseen environments. Fig.~\ref{fig:pre-exp-each} demonstrates the improvement of the success rate over the scene feature difference. Each point represents a distinct validation-unseen environment. The feature difference under each unseen environment is calculated as the mean of the L2-distance between the visual feature of all scenes from that environment and all scenes in the training environments. In general, most of the unseen environments gain improvement under the pre-exploration. We also find a trend that under the environment which has a larger feature difference, it can improve more under the pre-exploration. It shows that under more different environments, the pre-exploration can be more powerful which makes it practical to be more adaptive and generalized to real-life unseen environments. \\

\noindent\textbf{Qualitative Results~}
Fig.~\ref{fig:vis} demonstrates the visualization results of the navigation path without and with pre-exploration for the instruction \textit{``Walk out of the bathroom"}. Under the unseen environment, it is difficult to find out a path to get out of the unfamiliar bathroom, and as is shown in Fig.~\ref{fig:vis}(a), the model without pre-exploration is stuck inside. In contrast, with the knowledge learned during the pre-exploration phase, the model can successfully walk out of the bathroom and eventually achieve the final goal.

\section{Conclusion}
In this paper, we integrate counterfactual thinking into the vision-and-language navigation (VLN) task to solve the data scarcity problem. We realize counterfactual thinking via adversarial learning where we introduce an adversarial path sampler (APS) to only consider useful counterfactual conditions. The proposed APS is model-agnostic and proven effective in producing challenging but useful paths to boost the performances of different VLN models. 
Due to the power of reasoning, counterfactual thinking has gradually received attention in different fields. 
We believe that our adversarial training method is an effective solution to realize counterfactual thinking in general, which can possibly benefit more tasks.
\\ \\
\noindent\textbf{Acknowledgments.~}Research was sponsored by the U.S. Army Research Office and was accomplished under Contract Number W911NF-19-D-0001 for the Institute for Collaborative Biotechnologies. The views and conclusions contained in this document are those of the authors and should not be interpreted as representing the official policies, either expressed or implied, of the U.S. Government. The U.S. Government is authorized to reproduce and distribute reprints for Government purposes notwithstanding any copyright notation herein. 

\clearpage

\bibliographystyle{splncs04}
\bibliography{egbib}

\end{document}